\documentclass[journal]{IEEEtran}

\usepackage{verbatim}
\usepackage{amsfonts}
\usepackage{amssymb}
\usepackage{stfloats}
\usepackage{cite}
\usepackage{graphicx}
\usepackage{psfrag}
\usepackage{amsmath}
\usepackage{array}
\usepackage{epstopdf}
\usepackage{authblk}
\usepackage{graphicx} 
\usepackage{amsthm} 
\usepackage{lipsum}
\usepackage{verbatim} 
\usepackage{authblk}
\usepackage{mathtools}
\usepackage{cuted}
\usepackage{booktabs}
\usepackage{subfigure}

\newtheorem{theorem}{\bf Theorem}
\newtheorem{proposition}{\bf Proposition}
\newtheorem{lemma}{\bf Lemma}

\usepackage{cite}

\usepackage{amsmath}
\usepackage{mathrsfs}

\usepackage{algorithmic}

\usepackage{array}

\usepackage{url}
\usepackage{graphicx,amsmath,amssymb,amsfonts}
\usepackage{algorithmic,algorithm}

\usepackage{setspace}

\usepackage{xcolor}

\makeatletter

\newcommand{\Rmnum}[1]{\expandafter\@slowromancap\romannumeral #1@}
\makeatother

\usepackage{framed} 

\usepackage{cancel}

\begin{document}
% reference control
\bstctlcite{ref:BSTcontrol}

\title{Towards Communication-Learning Trade-off for Federated Learning at the Network Edge}

\author{Jianyang~Ren,~Wanli~Ni,\textit{~Graduate~Student~Member,~IEEE},~and~Hui~Tian,\textit{~Senior~Member,~IEEE}\vspace{-4.3mm} 
	  
		\thanks{This work was supported in part by the National Key R$\&$D Program of China under Grant No. 2020YFB1807801, and in part by the BUPT Innovation and Entrepreneurship Support Program under Grant No. 2022-YC-T010. \textit{(Corresponding author: Hui~Tian.)}}% <-this % stops a space
		\thanks{All authors are with the State Key Laboratory of Networking and Switching Technology, Beijing University of Posts and Telecommunications, Beijing 100876, China (e-mail: \{renjianyang, charleswall, tianhui\}@bupt.edu.cn).}
		
		}
	
\maketitle

\begin{abstract}
	In this letter, we study a wireless federated learning (FL) system where network pruning is applied to local users with limited resources.
	Although pruning is beneficial to reduce FL latency, it also deteriorates learning performance due to the information loss.
	Thus, a trade-off problem between communication and learning is raised.
	To address this challenge, we quantify the effects of network pruning and packet error on the learning performance by deriving the convergence rate of FL with a non-convex loss function.
	Then, closed-form solutions for pruning control and bandwidth allocation are proposed to minimize the weighted sum of FL latency and FL performance. 
	Finally, numerical results demonstrate that \romannumeral1) our proposed solution can outperform benchmarks in terms of cost reduction and accuracy guarantee, and \romannumeral2) a higher pruning rate would bring less communication overhead but also worsen FL accuracy, which is consistent with our theoretical analysis.
\end{abstract}
\vspace{-1mm}
\vspace{-1mm}
\vspace{-1mm}
\section{Introduction}
\vspace{-1mm}
\IEEEPARstart{T}{he} increasing popularity of mobile devices and the rapid development of smart applications have led to a significant increase in the amount of user data \cite{dinh2021federated}.
To make full use of these distributed datasets while protecting users privacy, federated learning (FL) has grabbed the limelight.
However, there are still some challenges when deploying FL in wireless networks.
On the one hand, the increasing number of model parameters induced by deep neural network (DNN) causes higher training and communication latency.
On the other hand, the packet error over unreliable wireless channels affects the accuracy of global aggregation.

Recently, many existing works have focused on communication efficient FL \cite{dinh2021federated, yang2020energy,luo2021cost,xu2021client,chen2021joint,ren2021joint}.
To reduce both FL latency and energy consumption, Dinh \textit{et. al} in \cite{dinh2021federated} derived closed-form solutions for resource allocation.
Yang \textit{et. al} in \cite{yang2020energy} proposed an iterative algorithm to minimize the total energy under FL latency constraint.
Likewise, Luo \textit{et. al} in \cite{luo2021cost} optimized the number of selected users to minimize the total cost while controlling the learning cost.
Focusing on the long-term performance, Xu \textit{et. al} in \cite{xu2021client} developed a joint client selection and bandwidth allocation algorithm under energy constraints. 
Furthermore, the effect of the packet error on FL convergence was derived in \cite{chen2021joint}.
Considering an edge computing-based FL system, Ren \textit{et. al} in \cite{ren2021joint} minimized the weighted sum of communication and learning cost.
However, in the above literature, the strong convexity assumption limits the applications of their schemes in DNN or other models with non-convex loss function.

With the increase of model complexity, the training latency of FL becomes critical for many time-sensitive scenarios such as autonomous driving and industrial control.
To make large-size models compatible to resource-limited devices, model compression has drawn great attention \cite{molchanov2019importance}.
Specifically,
Li \textit{et. al} in \cite{li2021talk} proposed a flexible compression scheme to balance FL latency and energy consumption.
However, the scheme designed in \cite{li2021talk} could only reduce communication latency.
To further reduce training latency, adaptive network pruning was adopted in \cite{jiang2020model} and \cite{liu2021adaptive}.
Jiang \textit{et. al} in \cite{jiang2020model} proposed a pruning-based FL scheme by adjusting the model size to reduce FL latency.
Furthermore, the effect of local pruning on FL convergence was analyzed in \cite{liu2021adaptive}.
Although the network pruning can reduce the model size of DNN, the resulting information loss also leads to the deterioration of learning behavior.
Thus, it is important to make a trade-off between communication and learning for achieving efficient FL. 

Overall, through optimizing resource allocation, the works in \cite{dinh2021federated, yang2020energy,luo2021cost,xu2021client,chen2021joint,ren2021joint} improved FL efficiency, but required clients with sufficient computing and storage capacity.
Although the model compression methods in \cite{li2021talk,jiang2020model,liu2021adaptive} can alleviate these issues, the effects of packet error and sample number on FL convergence are ignored.
Motivated by this background, we are committed to improve the communication-learning trade-off in an adaptive network pruning supported FL system.
The main contributions of this work include:
\begin{itemize}
  \item[1)] % 
  We derive the convergence upper bound of pruned FL, which reveals that both a higher pruning and packet error rate will worsen the convergence rate. Besides, users with more local training samples have greater impact on the achievable upper bound.
  \item[2)] 
  We formulate a non-convex problem to strike a balance between the communication and learning performance.
  Through decoupling problem, we provide closed-form solutions for network pruning rates and obtain the optimal bandwidth allocation with bisection method.
  \item[3)]Numerical simulations are conducted to validate the effectiveness of the proposed schemes. Experimental results show that our solutions can obtain better identification accuracy and lower total cost than benchmarks.
\end{itemize}
\vspace{-1mm}
\section{System Model}
\vspace{-1mm}
\label{system}

\begin{figure} [t!]
	\setlength{\abovecaptionskip}{0mm}
	\centering
	\includegraphics[width=3.65 in]{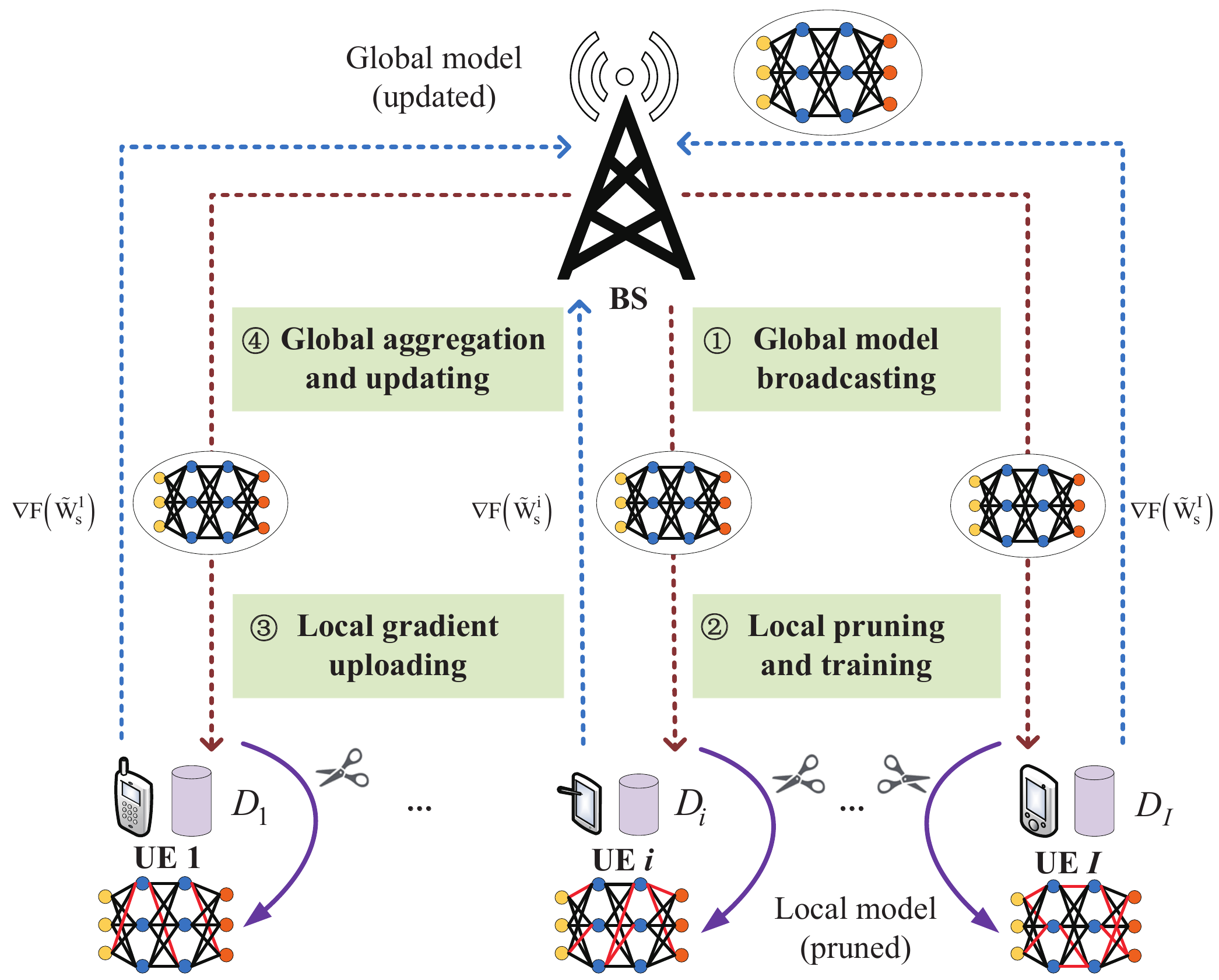}
	\caption{System model of pruned FL.}
	\label{system_model}
	\vspace{-5 mm}
\end{figure}

As shown in Fig. \ref{system_model}, we consider a FL system with network pruning, where there are one base station (BS) and $I$ user equipments (UEs), indexed by $\mathcal{I} = \{ 1,2, \ldots, I\}$.
%
%There are a certain amount of independent identically distributed (IID) samples in each UE's local dataset.
%
Let $\mathcal{D}_i$ denote the local dataset of UE $i$, where $K_i^{\rm max} = |\mathcal{D}_i|$ denotes the number of samples owned by UE $i$.
At the $s$-th round, the BS first broadcasts the latest global model $W_s$ to all UEs.
Upon reception, each UE prunes it to obtain a pruned model $\tilde{W}_s^{i}$ and starts its training with local dataset.
Then, all UEs uploads their gradients $\nabla{F}(\tilde{W}_s^i)\ (\forall i\in\mathcal{I})$ to the BS for global aggregation and model update. 

\subsection{Communication Latency Model}
The communication latency is defined as the time consumed by completing one communication round of pruned FL, which is composed of four parts as follows.

\subsubsection{Model Broadcasting Latency}
The achievable downlink transmission rate of UE $i$ is
\begin{equation}
	\setlength\abovedisplayskip{2pt}
	\setlength\belowdisplayskip{2pt}
	R_i^{\rm d}=B{\rm log}_2\left(1+p^{\rm d}h_i^{\rm d}/BN_0\right),
\end{equation}
where $B$ is the entire bandwidth at the BS, $p^{\rm d}$ is the transmit power, $h_i^{\rm d}$ is the downlink channel gain from the BS to UE $i$, and $N_0$ denotes the noise power spectral density.
The broadcast latency depends on the UE with the worst channel condition, which can be denoted by $t^{\rm d}=\max_{i}\left\{D_{\rm M}/R_i^{\rm d}\right\}$,
where $D_{\rm M}$ is the data size of the global model.
%the power spectral density of the Gaussian noise.

\subsubsection{Training Latency}
Upon receiving the global model, each UE prunes it to get a local model.
Let $\rho_i={D_{\rm P}^i}/{D_{\rm M}}$, denote the pruning rate at UE $i$, where $D_{\rm P}^i$ is the data size pruned by UE $i$.
Using federated stochastic gradient descent (FedSGD), the training latency for UE $i$ is specified as
\begin{equation}
	\setlength\abovedisplayskip{2pt}
	\setlength\belowdisplayskip{2pt}
	t_i^{\rm c}=\left(1-\rho_i\right){K_i}d^{\rm c}/{f_i},
\end{equation}
where $d^{\rm c}$ denotes the CPU cycles required to compute one sample at the BS \cite{liu2021adaptive}, $K_i$ out of $K_i^{\rm max}$ is the number of samples used by UE $i$ for local training and $f_i$ is the available CPU cycles per second at UE $i$.
%
%Since the total number of samples owned by UE $i$ is $K_i^{\rm max}$, so it is clear that $K_i$ should satisfy $0\le K_i\le K_i^{\rm max}$.
%
Compared with the training latency $t_i^{\rm c}$, the time consumed to prune the global model is short, thus neglected here.

\subsubsection{Gradient Uploading Latency}
In this letter, the frequency domain multiple access (FDMA) is adopted for local gradient uploading.
The achievable uplink transmission rate at UE $i$ is given by
\begin{equation}\label{uplink_trans_rate}
	\setlength\abovedisplayskip{2pt}
	\setlength\belowdisplayskip{2pt}
	R_i^{\rm u}=B_i{\rm log}_2\left(1+{p_ih_i^{\rm u}}/{B_iN_0}\right),
\end{equation}
where $B_i$ is the bandwidth allocated to UE $i$, $p_i$ is the maximum transmit power at UE $i$, and $h_i^{\rm u}$ is the uplink channel gain from UE $i$ to the BS.
%
%Since the total bandwidth is limited, we have $\sum_{i=1}^{I}B_i\le B$.
%
Due to the similarity in the data size of gradient and model, the gradient uploading latency at UE $i$ can be written as $t_i^{\rm u}=\left(1-\rho_i\right)D_{\rm M}/R_i^{\rm u}$.

\subsubsection{Global Aggregation Latency}
For simplicity, the global aggregation latency is given by a constant $t^{\rm a}$, which is affected by many factors, such as the hardware structure and time complexity of signal decoding.
To sum up, the FL latency for completing one communication round is given by
\begin{equation}\label{total_latency}
	\setlength\abovedisplayskip{2pt}
	\setlength\belowdisplayskip{-6pt}
	t=\max_{i\in\mathcal{I}}\left\{ t^{\rm d}+ t_i^{\rm c}+ t_i^{\rm u}+ t^{\rm a} \right\}.
\end{equation}
\subsection{Federated Learning Model}
To characterize the effect of packet error on the convergence rate of pruned FL, we first give the packet error rate at UE $i$ as
%\begin{equation}\
%$q_i={\rm E}_{h_i}\left[1-{\rm exp}\left({-m_0B_iN_0}/{p_ih_i^{\rm u}}\right)\right]$,
$q_i=1-{\rm exp}\left({-m_0B_iN_0}/{p_ih_i^{\rm u}}\right)$,
%\end{equation}
where $m_0$ is a waterfall threshold \cite{Xi2011ageneral}.
We assume that each local gradient is uploaded as a single packet without retransmissions scheme.
When the received local gradient contains errors, the BS will not aggregate it \cite{chen2021joint}.
Thus, the global gradient at the $s$-th round is
\begin{equation}
	\setlength\abovedisplayskip{2pt}
	\setlength\belowdisplayskip{2pt}
	g_s=\frac{\sum_{i=1}^{I}K_i\nabla F\left(\tilde{W}_s^i\right)C\left(\tilde{W}_s^i\right)}{\sum_{i=1}^IK_iC\left(\tilde{W}_s^i\right)},
\end{equation}
where $C\left(\cdot\right)$ is the indicator of packet error, expressed as
\begin{equation}
	\setlength\abovedisplayskip{2pt}
	\setlength\belowdisplayskip{2pt}
	C\left(\tilde{W}_s^i\right)=\left\{
	\begin{array}{rcl}
	1,  &  &{\rm with{\quad}probability{\quad}}1-q_i,\\
	0,  &  &{\rm with{\quad}probability{\quad}}q_i.
	\end{array}\right.
\end{equation}
With the obtained $g_s$, the global model is updated by $W_{s+1}=W_s-\eta g_s$, where $\eta$ is the learning rate.
\vspace{-1mm}
\section{Convergence and Problem Formulation} \label{convergence_analysis}
\vspace{-1mm}
\subsection{Convergence Analysis}
To facilitate analysis, we make the following assumptions.
\noindent\textbf{Assumption 1.} The loss function $F\left(\cdot\right)$ is $\beta$-smooth \cite{liu2021adaptive}:
\begin{equation}
	\setlength\abovedisplayskip{2pt}
	\setlength\belowdisplayskip{2pt}
	%\begin{aligned}
	F\left(V\right)\!\leq\!{F\left(W\right)}\!+\!\left<V-W,\nabla{F\left(W\right)}\right>\!+\!
	\frac{\beta}{2}{\left\|V-W\right\|^2}.
	%\end{aligned}
\end{equation} 

\noindent\textbf{Assumption 2.} The loss ${f_{ik}(\tilde{W}^i)}$ calculated on the $k$-th sample with the local model $\tilde{W}^i$ satisfies:
\begin{equation}
	\setlength\abovedisplayskip{2pt}
	\setlength\belowdisplayskip{2pt}
	\left\|\nabla{f_{ik}\left(\tilde{W}^i\right)}\right\|^2\leq{\xi_1+\xi_2\left\|\nabla{F\left({W}\right)}\right\|^2},
\end{equation} 
where $\xi_1$ and $\xi_2$ are two non-negative constants\cite{chen2021joint}.

\noindent
\textbf{Assumption 3.} The model weight is bounded by a non-negative constant $D$\cite{liu2021adaptive}, that is,
\begin{equation}
	\setlength\abovedisplayskip{2pt}
	\setlength\belowdisplayskip{2pt}
	%\mathbb{E}\left[\|W_s\|^2\right]\le D^2,
	\mathbb{E}\left[\|W\|^2\right]\le D^2.
\end{equation}
When Assumptions $1-3$ hold, the average $l_2$-norm of global gradient is used by us to evaluate the expected convergence rate of pruned FL, which is given in the following theorem.

%
%The convergence rate is given in the following theorem.
\begin{theorem}\label{theorem_convergence}
    Let $d=1-8\xi_2$. The expected convergence rate of pruned FL after $S$ communication rounds is given by
	\begin{align} \label{convergence_rate_norm}
		%\begin{aligned}
		\setlength\abovedisplayskip{1pt}
		\setlength\belowdisplayskip{1pt}
		&\frac{1}{S+1}\sum\nolimits_{s=0}^{S}{\!\mathbb{E} }\left[\left\|\nabla{F\left({W}_s\right)}\right\|^2\right]\leq \nonumber \\
		&\underbrace{\frac{F\!\left(W_0\right)\!-\!F\!\left(W^*\right)}{d\left(S+1\right)/2\beta}}_{\text{effect of initial model }}\!+\!\underbrace{\frac{8\xi_1}{dK}\!\sum_{i=1}^{I}\!K_i\overline{q}_i}_{\text{effect of packet error}}\!+\!\underbrace{\frac{2\beta^2ID^2}{dK^2}\!\sum_{i=1}^{I}\!K_i^2\overline{\rho}_i}_{\text{effect of network pruning}},
		%\end{aligned}
	\end{align}
    where $\overline{\rho}_i$ and $\overline{q}_i$ are the average pruning rate and the average packet error rate at UE $i$ during $S+1$ rounds, respectively.
\end{theorem}

\begin{IEEEproof}
     Please see Appendix A.
\end{IEEEproof}

Based on Theorem \ref{theorem_convergence}, we can find that the average $l_2$-norm of global gradient is bounded by the sum of three terms.
The first term is affected by the gap between the initial model and the optimal model, which converges to zero as $S$ goes to infinity.
The second term reflects the effect of average packet error rate.
A higher average packet error rate $\bar{q}_i$ will result in a larger upper bound of the average $l_2$-norm of global gradient. %, that is the lower convergence rate.
This result is consistent with the actual situation, since a higher packet error rate means a higher probability that the local gradient may not be applied to the global aggregation, which reduces the learning performance.
The third term is related to the average local pruning rate $\bar{\rho}_i$.
A higher pruning rate causes a greater deviation between the pruned model and global model, thus increases the upper bound.
In addition, we can find that the number of local training samples used by UEs also affects the upper bound.
To increase the convergence rate, reducing the average packet error rate and pruning rate for UEs that use more samples for local learning is more beneficial.

Since the effects of packet error and network pruning can not be mitigated by only increasing communication rounds $S$, we focus on reducing them through optimization.

It can be found from (\ref{convergence_rate_norm}) that optimizing the average $l_2$-norm of global gradient is equivalent to optimize the one-round convergence upper bound, which is defined by
    \begin{equation}\label{convergence_upper_bound}
    	\setlength\abovedisplayskip{3pt}
    	\setlength\belowdisplayskip{3pt}
	    \gamma=\psi+m\sum\nolimits_{i=1}^{I}K_i\left(q_i+K_i\rho_i\right),
    \end{equation} 
where $m=\max\{8\xi_1/dK,\ 2\beta^2ID^2/dK^2\}$ and $\psi={2\beta}\left(F\left(W_0\right)-F\left(W^*\right)\right)/d(S+1)$.

\subsection{Problem Formulation}
Both FL latency and learning performance are important to the practical implementation of pruned FL at the network edge, thus it is desirable to minimize the FL latency in (\ref{total_latency}) while reducing the upper bound in (\ref{convergence_upper_bound}).
However, it is difficult to minimize two metrics at the same time.
For example, increasing the local pruning rates is helpful to reduce FL latency, but it inevitably degrades the convergence rate.
Therefore, to strike a trade-off between the communication and learning performance, 
we formulated a weighted sum optimization problem, which is given by
\begin{subequations}\label{total_cost_minimization}
	\setlength\abovedisplayskip{2pt}
	\setlength\belowdisplayskip{2pt}
	\begin{eqnarray}
		%(\mathcal{P}0): 
		\label{total_cost_minimization_objective}
		&\min \limits_{ \boldsymbol{\rho}, \mathbf{B}} & \left(1-\lambda\right)t+\lambda\gamma \\
		\label{total_cost_minimization_prune_constraint}
		&{\rm s.t.}&0 \le \rho_i \le \rho_i^{\rm max},\ \forall i,\\
		\label{total_cost_minimization_band_constraint1}
		&{}&\sum\nolimits_{i=1}^{I}B_i\leq B, \\
		\label{total_cost_minimization_band_constraint2}
		&{}&B_i\geq 0,\ \forall i,  
	\end{eqnarray}
\end{subequations}
where $\lambda \in [0,1]$ is a weight to balance two metrics depending on scenario and their magnitude difference, $\boldsymbol{\rho}=[\rho_1,\rho_2,\ldots, \rho_I]^T$ and $\mathbf{B}=[B_1,B_2,\ldots,B_I]^T$ denote the network pruning and bandwidth allocation vectors, respectively. 
In constraint (\ref{total_cost_minimization_prune_constraint}), $\rho_i^{\rm max}$ denotes the maximum pruning rate at UE $i$, which depends on the acceptable information loss.
Since the total bandwidth is limited, we have constraint (\ref{total_cost_minimization_band_constraint1}).
Constraint (\ref{total_cost_minimization_band_constraint2}) restricts the bandwidth allocation to UEs.
Due to the close coupling of the optimization variables in (\ref{total_cost_minimization_objective}), problem (\ref{total_cost_minimization}) is a non-convex optimization problem, which is hard to solve directly.
In the following, by introducing an auxiliary variable to transform (\ref{total_cost_minimization}) equivalently, we provide closed-form solutions to find the design of UEs' local pruning rates and bandwidth allocation.

\vspace{-1mm}
\section{Proposed Solution}
\vspace{-1mm}
Before solving problem (\ref{total_cost_minimization}), we first rearrange (\ref{total_cost_minimization_objective}) as
\begin{equation}\label{total_cost_minimization_objective_new}
	\setlength\abovedisplayskip{3pt}
	\setlength\belowdisplayskip{3pt}
	%\begin{aligned}
	F\left(\boldsymbol{\rho},\mathbf{B}\right)=\left(1-\lambda\right)\max \nolimits_{i}\left\{t_i^{\rm c}+t_i^{\rm u}\right\}+\upsilon+\lambda\left(\gamma-\psi\right), %m\sum\nolimits_{i=1}^I K_i\left(q_i+K_i\rho_i\right),
    %\end{aligned}
\end{equation} 
where $\upsilon=(1-\lambda) (t^{\rm d}+t^{\rm a})+\lambda\psi$ is independent with all optimization variables.
By introducing an auxiliary variable $\tilde{t}$, problem (\ref{total_cost_minimization}) can be equivalently transformed to
\begin{subequations}\label{total_cost_minimization_aux}
	\begin{eqnarray}
		%(\mathcal{P}1): 
		\label{total_cost_minimization_objective_aux}
		&\min \limits_{ \boldsymbol{\rho}, \mathbf{B}, \tilde{t}} & \left(1-\lambda\right)\tilde{t}+\lambda m\sum\nolimits_{i=1}^I K_i\left(q_i+K_i\rho_i\right)\\%+\upsilon \\
		\label{total_cost_min_aux_time_constraint}
		&{\rm s.t.}&t_i^{\rm c}+t_i^{\rm u}\le \tilde{t},\ \forall i, \\
		&{}&(\ref{total_cost_minimization_prune_constraint})-(\ref{total_cost_minimization_band_constraint2}),
	\end{eqnarray}
\end{subequations}
where (\ref{total_cost_minimization_objective_aux}) is our considered total cost, the weighted sum of FL latency and learning cost.
To solve this problem, we decouple it into two sub-problems and derive the corresponding closed-form solutions.

\subsection{Optimization of Pruning Rates}
Given the bandwidth allocation scheme $\mathbf{B}$, problem (\ref{total_cost_minimization_aux}) can be rewritten as
\begin{subequations}\label{latency_purning_optimization}
	\setlength\abovedisplayskip{2pt}
	\setlength\belowdisplayskip{2pt}
	\begin{eqnarray}
		%(\mathcal{P}2): 
		\label{latency_purning_opt_objective}
		&\min \limits_{ \boldsymbol{\rho}, \tilde{t}} & \left(1-\lambda\right)\tilde{t}+\lambda m\sum\nolimits_{i=1}^IK_i^2\rho_i \\
		&{s.t.}&(\ref{total_cost_minimization_prune_constraint})\ \text{and}\ (\ref{total_cost_min_aux_time_constraint}), 
	\end{eqnarray}
\end{subequations}
which is a linear programming problem of $\boldsymbol{\rho}$ and $\tilde{t}$.
From (\ref{latency_purning_opt_objective}), it is always efficient to utilize the minimal pruning rates, which can be derived from (\ref{total_cost_min_aux_time_constraint}) as
\begin{equation}\label{local_pruning_rate_min}
	\setlength\abovedisplayskip{3pt}
	\setlength\belowdisplayskip{3pt}
	\rho_i^{\rm min}\left(\tilde{t}\right) = \max\left\{1-\frac{\tilde{t}}{D_{\rm M}/R_i^{\rm u}+K_id^{\rm c}/f_i},0\right\},\ \forall i.
\end{equation}

Substituting $(\ref{local_pruning_rate_min})$ into problem (\ref{latency_purning_optimization}) yields:
\begin{subequations}\label{latency_optimization}
	\setlength\abovedisplayskip{3pt}
	\setlength\belowdisplayskip{3pt}
	\begin{eqnarray}
		\label{latency_opt_objective}
		&\min \limits_{ \tilde{t}} & \left(1-\lambda\right)\tilde{t}+\lambda m\sum\nolimits_{i=1}^IK_i^2\rho_i^{\rm min}\left(\tilde{t}\right) \\
		\label{latency_opt_time_constraint}
		&{\rm s.t.}&\tilde{t}^{\rm min} \le \tilde{t} \le \tilde{t}^{\rm max},
	\end{eqnarray}
\end{subequations}
where
$
    \tilde{t}^{\rm min}=\max_{i\in \mathcal{I}} \{ (\frac{D_{\rm M}}{R_i^{\rm u}}+\frac{K_id^{\rm c}}{f_i} )\left(1-\rho_i^{\rm max}\right) \}
$
and
$
	\tilde{t}^{\rm max}=\max_{i\in \mathcal{I}} \{\frac{D_{\rm M}}{R_i^{\rm u}}+\frac{K_id^{\rm c}}{f_i} \}.
$
The objective function (\ref{latency_opt_objective}) is a piece-wise linear function, where the required FL latency with no pruning $t_i^{\rm np}=D_{\rm M}/R_i^{\rm u}+K_id^{\rm c}/f_i\ \left(\forall i\in\mathcal{I}\right)$ are breakpoints.
Without loss of generality, we assume that they are sorted in a non-increasing order, i.e. $t_1^{\rm np}\ge t_2^{\rm np}\ge \ldots\ge t_I^{\rm np}$. %i.e.$\frac{D_{\rm M}}{R_1^{\rm u}}+\frac{K_1d^{\rm c}}{f_1}\ge\frac{D_{\rm M}}{R_2^{\rm u}}+\frac{K_1d^{\rm c}}{f_2}\ge \ldots\ge\frac{D_{\rm M}}{R_I^{\rm u}}+\frac{K_1d^{\rm c}}{f_I}$.

Further, we introduce an auxiliary variable $i_1$.
If $t_I^{\rm np} \ge \tilde{t}^{\rm min}$, then set $i_1=I$.
Otherwise we search for $i_1$ that satisfies $t_{i_1+1}^{\rm np}\le\tilde{t}^{\rm min}$ and $t_{i_1}^{\rm np}\ge\tilde{t}^{\rm min}$.
To this end, the closed-form optimal solution $\tilde{t}^*$ for the sub-problem (\ref{latency_optimization}) is given in the following proposition.
\begin{proposition}\label{proposition_1}
	The optimal solution $\tilde{t}^*$ can be expressed as
     \begin{equation}\label{t_opt}
     	\setlength\abovedisplayskip{2pt}
     	\setlength\belowdisplayskip{2pt}
     	\tilde{t}^*=\left\{
     	\begin{array}{lll}
     		\!\tilde{t}^{\rm min},&&\!1\!-\!\lambda-\!\lambda m\sum\limits_{i=1}^{i_1}\!\frac{K_i^2}{D_{\rm M}/R_i^{\rm u}+K_id^{\rm c}/f_i}\!\ge\! 0,\\
     		%\frac{D_{\rm M}}{R_{i_2}^{\rm u}}+\frac{K_1d^{\rm c}}{f_{i_2}}, &&otherwise,
     		\!t_{i_2}^{\rm np},&&{\rm otherwise},
     	\end{array}\right.
     \end{equation}
where $t_{i_2}^{\rm np}$ is the inflection point of function (\ref{latency_opt_objective}).
With calculated $\tilde{t}^*$, we can calculate the optimal local pruning rate $\rho_i^*$ of UE $i$ according to (\ref{local_pruning_rate_min}).
\end{proposition}

\subsection{Optimization of Bandwidth Allocation}
Given $\boldsymbol{\rho}$ and $\tilde{t}$, problem (\ref{total_cost_minimization_aux}) can be simplified as
\begin{subequations}\label{b_optimization}
	\setlength\abovedisplayskip{2pt}
	\setlength\belowdisplayskip{2pt}
	\begin{eqnarray}
		%(\mathcal{P}3):
		\label{b_optimization_objective}
		&\min \limits_{ \mathbf{B}} & \lambda m\sum\nolimits_{i=1}^IK_iq_i\\
		&{s.t.}&(\ref{total_cost_minimization_band_constraint1}), 
		(\ref{total_cost_minimization_band_constraint2})\ \rm{and}\ (\ref{total_cost_min_aux_time_constraint}).  
	\end{eqnarray}
\end{subequations}
We temporarily remove the constraint (\ref{total_cost_minimization_band_constraint1}) and use it to verify the feasibility of the solution to the simplified problem latter. 
Without constraint (\ref{total_cost_minimization_band_constraint1}), %the optimization variables in problem (\ref{b_optimization}) are now independent of each other. 
problem (\ref{b_optimization}) can be decoupled to $I$ independent sub-problems, each related to one UE.
The bandwidth allocation subproblem for UE $i$ is given by
\begin{subequations}\label{b_opt_decomposition}
	\begin{eqnarray}
		\label{b_opt_decomposition_objective}
		&\min \limits_{ B_i } & m \lambda K_i q_i\\
		%\left(1-{\rm exp}\left(\frac{-m_0B_iN_0}{p_ih_i}\right)\right)\\
		\label{b_opt_decomposition_time_constraint}
		&{\rm s.t.}&R_i^{\rm u}\ge \frac{\left(1-\rho_i\right)D_{\rm M}}{\tilde{t}-\left(1-\rho_i\right)K_id^{\rm c}/f_i},\\
		\label{b_opt_decomposition_band_constraint}
		&{}&B_i\ge 0, 
	\end{eqnarray}
\end{subequations}
where (\ref{b_opt_decomposition_time_constraint}) is transformed from (\ref{total_cost_min_aux_time_constraint}).
Rewrite $q_i$ and $R_i^{\rm u}$ into the form of functions, as $q_i\left(B_i\right)$ and $R_i^{\rm u}\left(B_i\right)$, whose important properties are shown in the following lemma.

\begin{lemma}\label{lemma_1}
	Both $q_i\left(B_i\right)$ and $R_i^{\rm u}\left(B_i\right)$ are monotonically increasing functions of variable $B_i$.
\end{lemma}

\begin{IEEEproof}
	The first order derivative of $q_i\left(B_i\right)$ and $R_i^{\rm u}\left(B_i\right)$ are $q_i^{\prime}\left(B_i\right)=\frac{m_0N_0}{p_ih_i^{\rm u}}{\rm exp}(-\frac{m_0B_iN_0}{p_ih_i^{\rm u}})\textgreater 0$ and ${R_i^{\rm{u}}}^{\prime}\left(B_i\right)={\rm log}_2(1+\frac{p_ih_i^{\rm u}}{B_iN_0})-\frac{p_ih_i^{\rm u}}{\left(B_iN_0+p_ih_i^{\rm u}\right){\rm ln}2}$, respectively.
	It is hard to give the range of ${R_i^{\rm{u}}}^{\prime}\left(B_i\right)$, thus we calculate
	${R_i^{\rm{u}}}^{\prime\prime}\left(B_i\right)=\frac{p_ih_i^{\rm u}\left(\theta_2-\theta_1\right)}{{\theta_1\theta_2 \rm ln}2}\textless 0$,
	where $\theta_1=B_i^2N_0+2B_ip_ih_i^{\rm u}+{p_i^2{h_i^{\rm u}}^2}/{N_0}$ and $\theta_2=B_i^2N_0+B_ip_ih_i^{\rm u}$.
	%
	%Due to $\theta_2-\theta_1 \textless 0$, ${R_i^{\rm{u}}}^{\prime}\left(B_i\right)$ is a monotonically decreasing function.
	%
	Finally, with ${R_i^{\rm{u}}}^{\prime\prime}\left(B_i\right)\textless 0$ and $\lim\nolimits_{B_i\to \infty}{R_i^{\rm{u}}}^{\prime}\left(B_i\right)=0$, we have ${R_i^{\rm{u}}}^{\prime}\left(B_i\right) \textgreater 0$.
	% which completes the proof.
\end{IEEEproof}

Based on Lemma \ref{lemma_1}, the optimal bandwidth allocation for UE $i$ equals to the minimum bandwidth under all constraints of problem (\ref{b_opt_decomposition}), which satisfies
\begin{equation}\label{b_opt_solution}
	\setlength\abovedisplayskip{2pt}
	\setlength\belowdisplayskip{2pt}
	R_i^{\rm u}\left(B_i^*\right)=\frac{\left(1-\rho_i\right)D_{\rm M}}{\tilde{t}-\left(1-\rho_i\right)K_id^{\rm c}/f_i},
\end{equation} 
With the monotonicity of $R_i^{\rm u}$, $B_i^*$ can be obtained by bisection method and its optimality is given in the following lemma. %\cite{wen2021adaptive}.
%
%Then, the optimality of $\mathbf{B}^*$ is given in the following lemma.
\begin{lemma}\label{lemma_2}
	Since $\sum_{i=1}^IB_i^*\le B$ always holds, $\mathbf{B}^*$ is the optimal solution of problem (\ref{b_optimization}).
\end{lemma}
\begin{IEEEproof}
	We denote the bandwidth allocation in the $n$-th iteration as $\mathbf{B}^{(n)}=\{B_1^{(n)}\!,B_2^{(n)}\!,\ldots\!,B_I^{(n)}\}$.
	In the $(n+1)$-th iteration, $\mathbf{B}^{(n)}$ and $B_i^{(n)}$ are still feasible for problem (\ref{b_optimization}) and (\ref{b_opt_decomposition}), respectively.
    Since the latest bandwidth allocation $B_i^{(n+1)}$ is the minimal feasible bandwidth under all constraints of problem (\ref{b_opt_decomposition}), it is obvious that $B_i^{(n+1)}\le B_i^{\rm (n)},\ \forall i$.
	Thus, we can get $\sum\nolimits_{i=1}^I B_i^{(n+1)}\le \sum\nolimits_{i=1}^I B_i^{(n)}\le B$.
\end{IEEEproof}
Therefore, our algorithm 
%for solving problem (\ref{total_cost_minimization_aux})
is summarized in Algorithm \ref{overall_algorithm}. 
The complexity of one iteration is $\mathcal{O}\left(I{\rm log}_{2}\left(1/\delta \right)\right)$, where $\delta$ is the convergence accuracy of bisection method.
For implementation, $\lambda$ is first searched based on FL latency requirement. Then, the BS runs the algorithm with the collected information, such as UEs' sample number, CPU frequency and channel gain.

\begin{algorithm}[t!]
	\caption{Overall Algorithm for Solving Problem (\ref{total_cost_minimization_aux})}
	\label{overall_algorithm}
	\begin{algorithmic}[1]
		\renewcommand{\algorithmicrequire}{\textbf{Initialize}}
		\renewcommand{\algorithmicensure}{\textbf{Output}}
		\STATE \textbf{Initialize} $\lambda$, $K_i$, $D_{\rm M}$, $d^{\rm c}$, $m$, $f_i$.
		\STATE \textbf{Set} the iteration number $j=1$. 
		\REPEAT
		\STATE Given $\mathbf{B}^{(j)}$, compute $\tilde{t}^{*(j)}$ using (\ref{t_opt}).
		\STATE With obtained $\tilde{t}^{*(j)}$, compute $\boldsymbol{\rho}^{(j)}=\{\rho_i^{*(j)}\}$ using (\ref{local_pruning_rate_min}).
		\STATE Given $\boldsymbol{\rho}^{(j)}$, solve (\ref{b_opt_solution}) with bisection method to obtain $\mathbf{B}^{(j+1)}=\{B_i^{*(j+1)}\}$.
		\STATE $j\leftarrow j+1$.
		\UNTIL the termination criterion is reached.
		\STATE \textbf{Output} the convergent solution $\{\boldsymbol{\rho}^*,\mathbf{B}^*,\tilde{t}^{*}\}$. 
	\end{algorithmic}
\vspace{-1mm}
\end{algorithm} 

\vspace{-1mm}
\section{Simulation Results}
\vspace{-1mm}
We examine our theoretical results in a pruned FL system with $I=5$ users. 
Each UE's CPU cycle is set as $5 {\rm \ GHz}$ and other parameters are shown in Table \ref{paramater_table}.
The shallow neural network ($\eta=10^{-3}$) and a DNN ($\eta=10^{-4}$) are trained on the MNIST and Fashion-MNIST dataset, respectively.\footnote{The shallow neural network has one hidden layer of 60 neurons. The DNN has two hidden layers of 60 and 20 neurons, respectively.}
Cross-entropy is adopted as the loss function.

Along with the exhaustive search with exponential complexity, the following schemes are also considered as benchmarks.
\begin{itemize}
    \item Greedy bandwidth allocation (GBA): Bandwidth allocation is proportional to the reciprocal of channel gain.\ % to mitigate the straggler issue of synchronous FL.
    \item Fixed pruning rate (FPR): The pruning rates for users are preset as constants, e.g., $\rho_i \in \{ 0, 0.35, 0.7 \}, \forall i$.
    \item Ideal FL: All local models are not pruned. Meanwhile, the packet error rates are assumed to be zero.
\end{itemize}

\begin{table}[t!]
	\setlength{\abovecaptionskip}{-0.3mm}
	\caption{Parameter Settings}\label{paramater_table}
	\centering
	\begin {tabular}{|c|c|c|c|}
	%\toprule
	\hline
	\textbf{Parameter}&\textbf{Value}&\textbf{Parameter}&\textbf{Value}\\
	%\midrule
	\hline
	$p_i\ (\forall i)$&23 dBm&$\rho_i^{\rm max}\ (\forall i)$&0.7\\
	\hline
	$D_{\rm M}$&1.6 Mbit&$B$&15 MHz\\
	\hline
	$m_0$&0.023 dB&$\lambda$&0.0004\\
	\hline
	$N_0$&-174 dBm/Hz&$d^{\rm c}$&0.168 GHz \\
	\hline
	\multicolumn{2}{|c|}{$K_i\in\{30,40,50\}, \forall i\in\mathcal{I}$}&local SGD step &1 \\
	%\bottomrule
	\hline
\end{tabular}
\vspace{-6 mm}
\end{table}

\begin{figure*}[t!]
	\setlength{\abovecaptionskip}{-0.5mm}
	\centering
		\begin{minipage}[t]{0.19\textwidth}
		\centering
		\includegraphics[width=1.36 in]{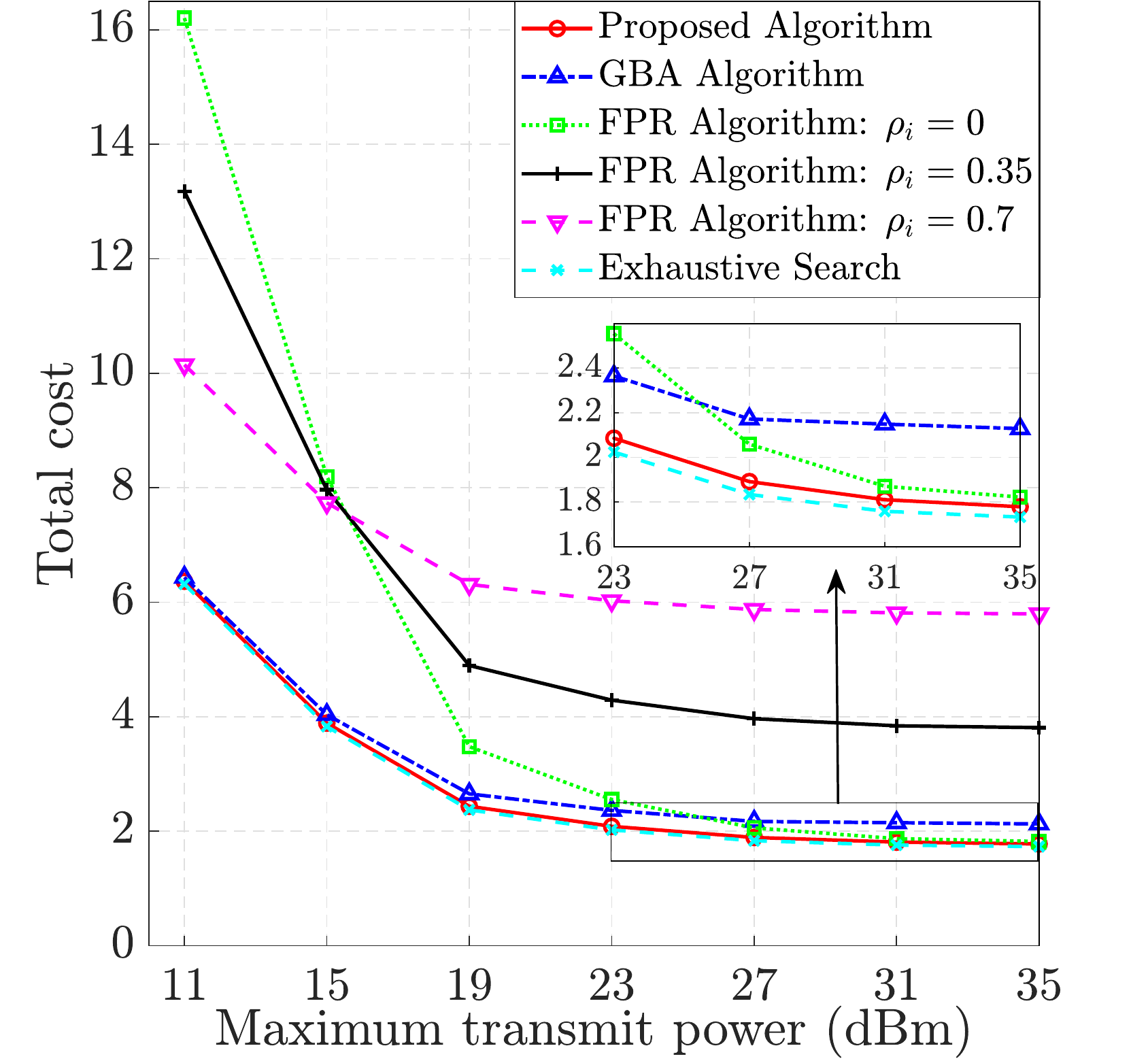}
		\caption{Total cost vs. $p_i$.}
		\label{transmit_power}
		\end{minipage}
	   \begin{minipage}[t]{0.19\textwidth}
		\centering
		\includegraphics[width=1.36 in]{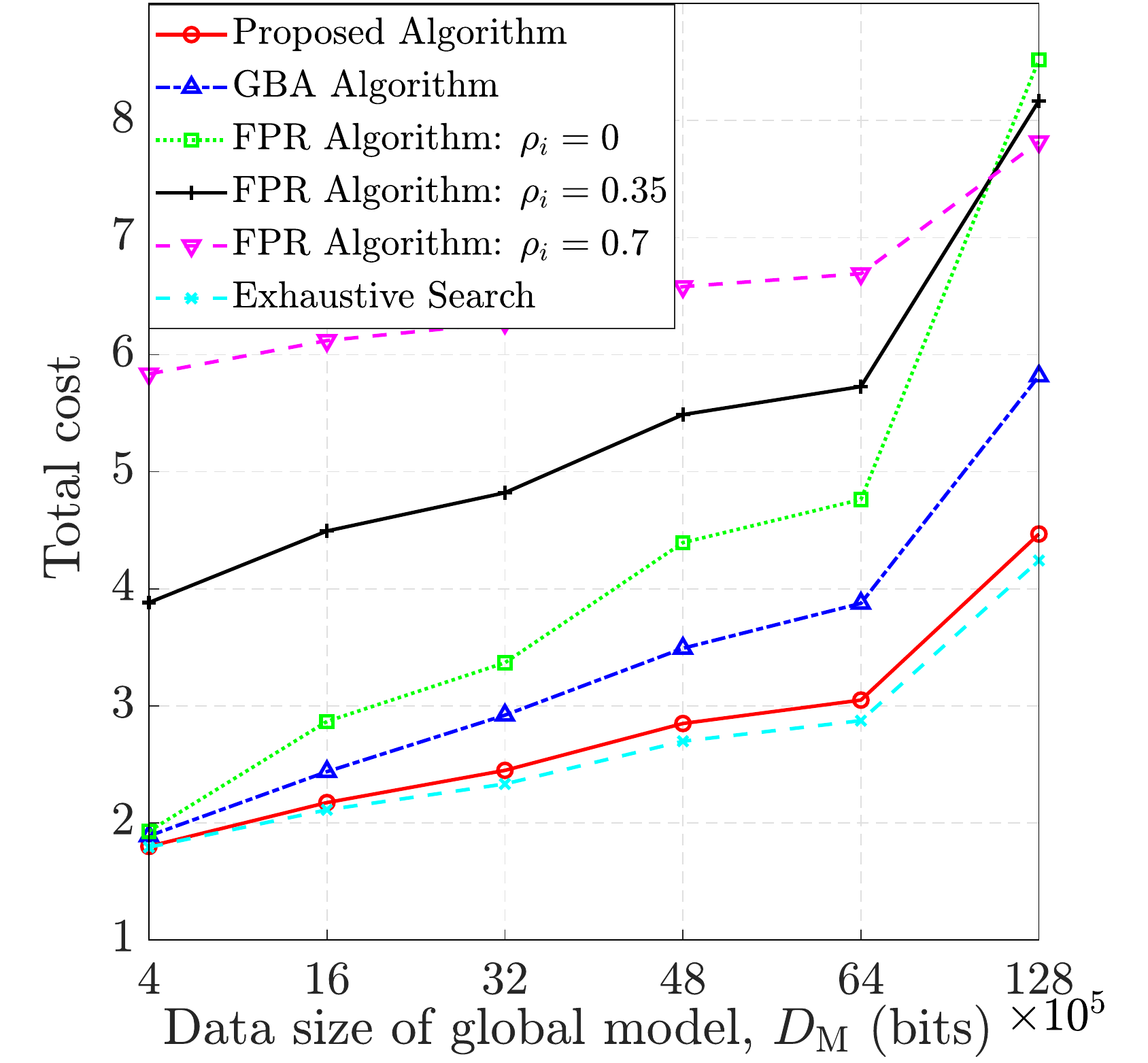}
		\caption{Total cost vs. $D_{\rm M}$.}
		\label{data_size}
	    \end{minipage}
    	\begin{minipage}[t]{0.19\textwidth}
    	\centering
    	\includegraphics[width=1.365 in]{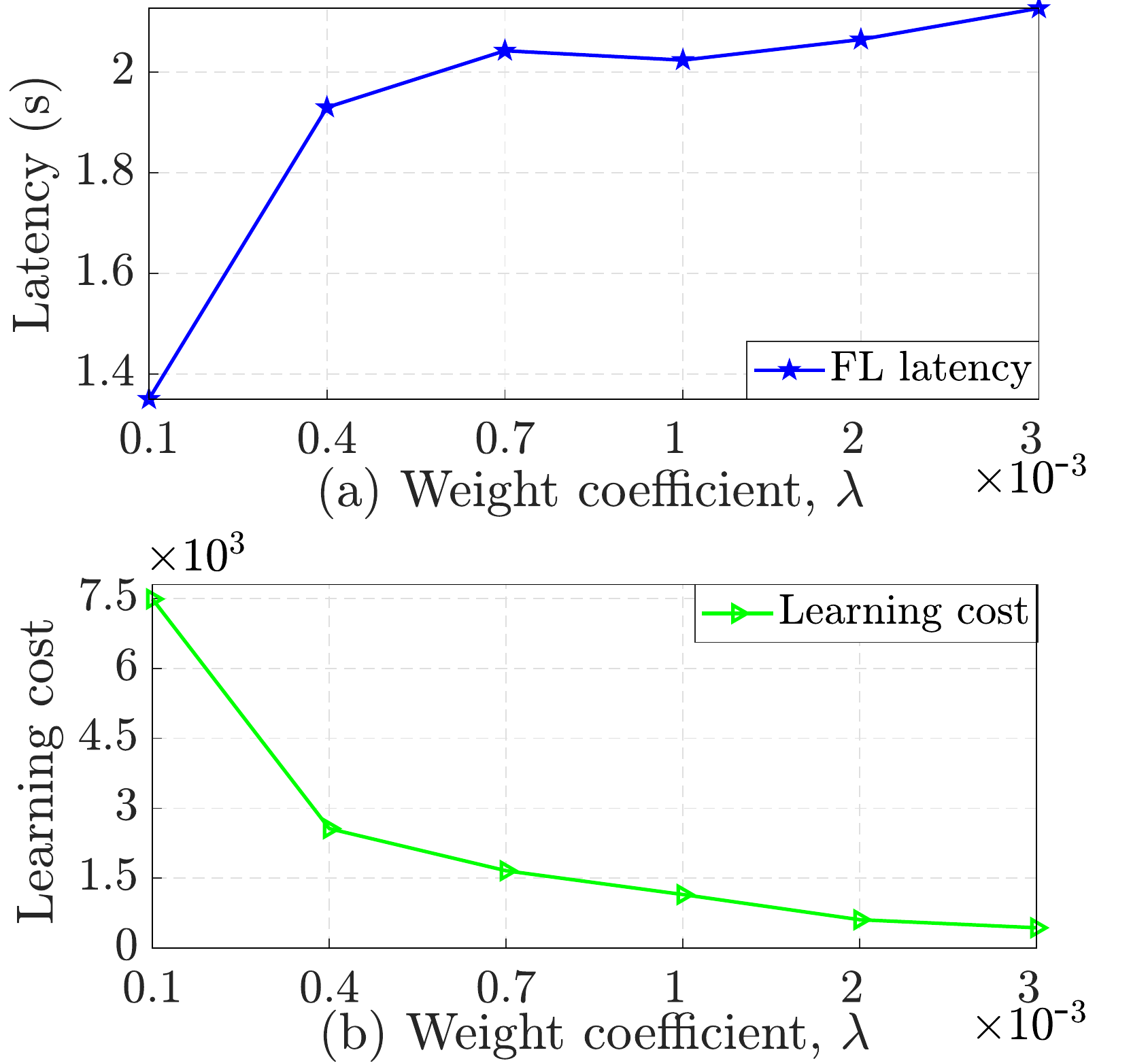}
    	\caption{Impact of $\lambda$.}
    	\label{weight}
    	\end{minipage}
    	\begin{minipage}[t]{0.19\textwidth}
    	\centering
    	\includegraphics[width=1.36 in]{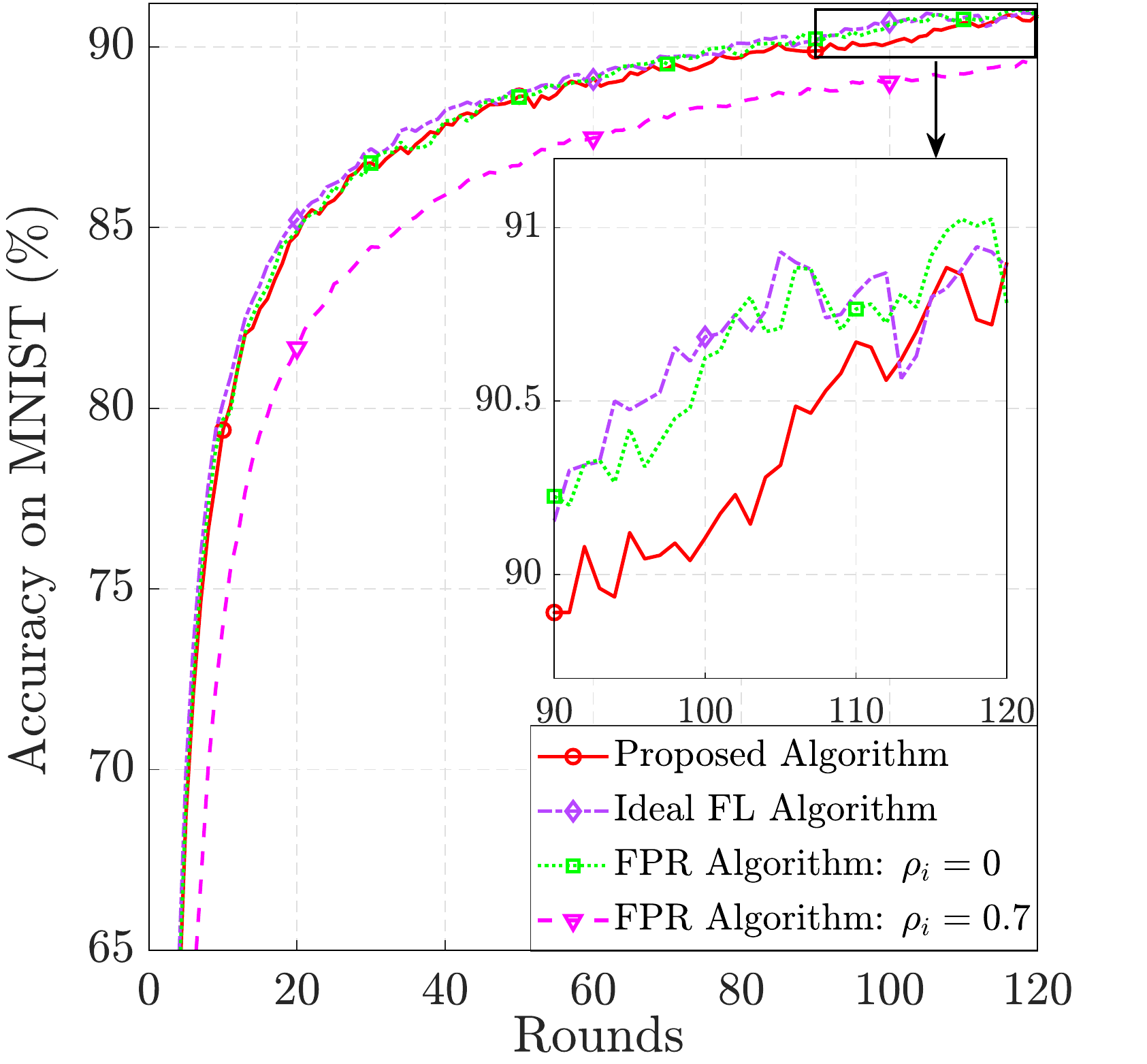}
    	\caption{Accuracy comparison.}
    	\label{accuracy}
    	\end{minipage}
    	\begin{minipage}[t]{0.19\textwidth}
    	\centering
    	\includegraphics[width=1.36 in]{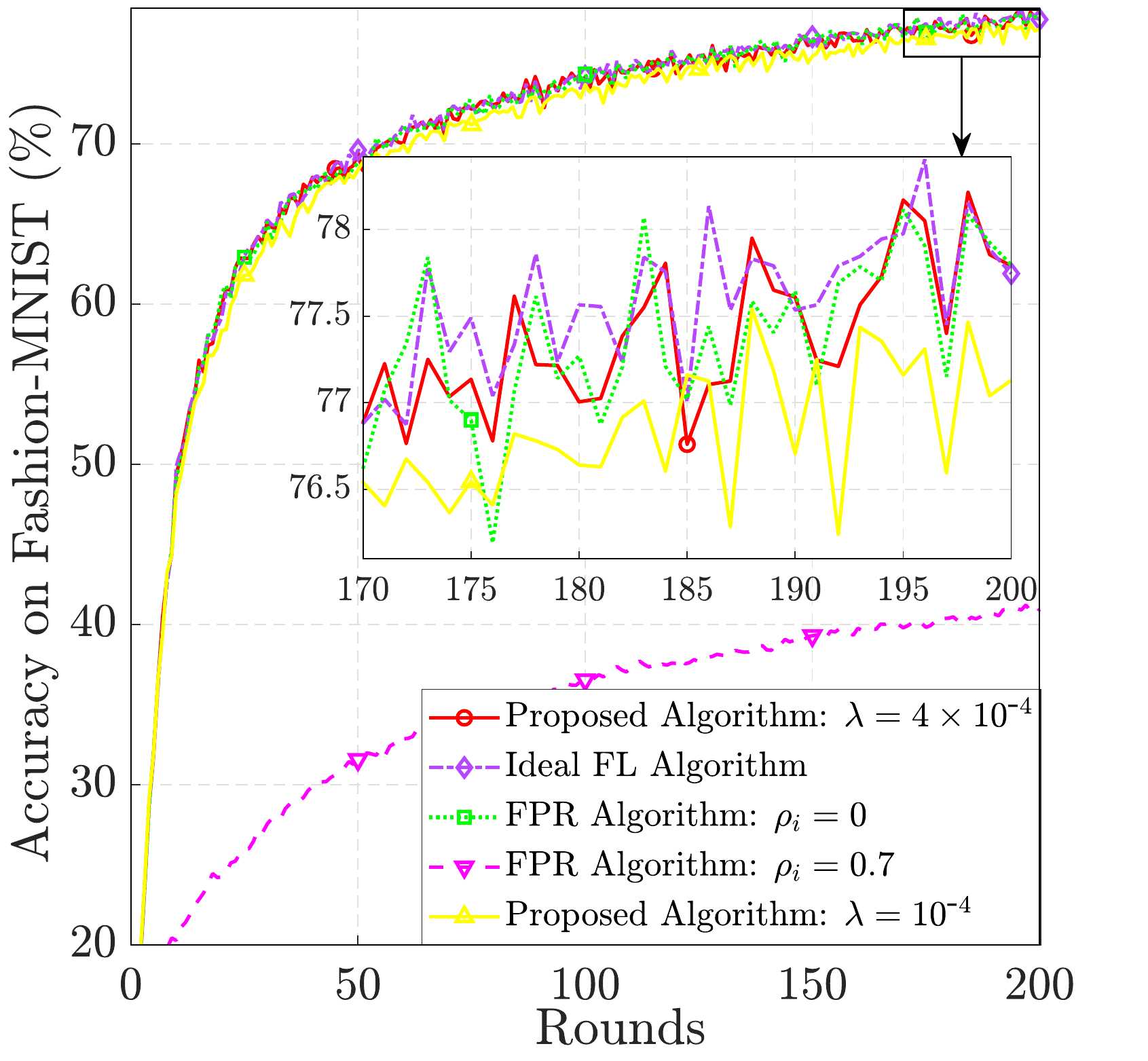}
    	\caption{Accuracy comparison.}
    	\label{acc_fashion}
    	\end{minipage}
    \vspace{-5mm}
\end{figure*}
Fig. \ref{transmit_power} presents the impact of the maximum transmit power on the total cost.
The total cost decreases as the maximum transmit power at UEs grows.
That is because higher transmit power is helpful to reduce FL latency and learning cost.
When UEs' maximum transmit power is low, higher pruning rates help to reduce the total cost. %which shows latency is the dominant factor in the total cost at the moment. 
With the increased maximum transmit power, the importance of FL latency relative to learning cost reduces, so reducing pruning rates brings benefits.
From this figure, we observe that the proposed solution outperforms baselines such as GBA and FPR algorithms, and is close to the exhaustive search.
Then we investigate the impact of the data size of global model in Fig. \ref{data_size}.
In the case of low data size, the proposed solution, GBA algorithm and FPR algorithm with $\rho_i=0$ show similar performance, because the total bandwidth is sufficient to transmit the low data size local gradient even with no pruning.
As data size increases, the proposed solution is always close to exhaustive search and enjoys a significant performance gain than others.
Finally, the impact of $\lambda$ is studied in Fig. \ref{weight}. As $\lambda$ grows, the system is more concentrated on the minimization of learning cost.
To this end, FL latency increases while learning cost decreases.
%

%\subsection{Learning Performance}
%
Fig. \ref{accuracy} shows the test accuracy of training a shallow neural network on MNIST dataset.
The results are averaged over twenty simulations.
Without considering pruning and packet error, the accuracy of the ideal FL is usually the highest.
The FPR algorithm with $\rho_i=0$ shows a slightly lower convergence rate than that of the ideal FL algorithm because of the packet error.
Note that the test accuracy of our proposed solution is usually slightly lower than that of ideal FL and FPR algorithm with $\rho_i=0$, because it prunes the local models to reduce FL latency.
Due to the high pruning rates, the accuracy achieved by the FPR algorithm with $\rho_i=0.7$ is the lowest.
In Fig. \ref{acc_fashion}, we present the test accuracy of training a DNN on Fashion-MNIST dataset. 
We notice that the proposed algorithm is close to the ideal FL, while the achievable performance of FPR algorithm with $\rho_i = 0.7$ is extremely lower than others. 
Besides, a smaller $\lambda$ leads to lower accuracy of our algorithm.

\vspace{-1mm}
\section{Conclusion}
\vspace{-1mm}
In this letter, a network pruning supported wireless FL system was studied.
We first theoretically analyzed the effect of the packet error and local pruning rates on the FL convergence upper bound. %in a wireless FL system supported by network pruning.
By capturing the trade-off between communication and learning, the closed-form solutions were derived to solve the formulated non-convex problem for total cost minimization.
Numerical results validated our theoretical analysis and demonstrated that our proposed scheme can reduce the total cost while maintaining the learning performance.

\section*{Appendix:Proof of Theorem 1}
The update function of global model can be rewritten as  $W_{s+1}=W_s-\eta\left(\nabla{F\left({W}_s\right)}-o\right)$, where $o=\nabla{F\left({W}_s\right)}-g_s$.
Let $\eta=\frac{1}{\beta}$ and further take the expectation of both sides of assumption 1, we have:
	\begin{align}\label{A_2}
		\setlength\abovedisplayskip{2pt}
		\setlength\belowdisplayskip{2pt}
		%\begin{aligned}
		\mathbb{E}\left[F\left(W_{s+1}\right)\right]\!\le\!F\left(W_s\right)\!-\!\frac{\mathbb{E}\left[\|\!\nabla F\!\left({W}_s\!\right)\!\|^2\right]}{2\beta}\!+\!\frac{\mathbb{E}\left[\|o\|^2\right]}{2\beta}.
	\end{align} 

Note that  $K=\sum_{i=1}^IK_i$, we further derive $\mathbb{E}\left[\|o\|^2\right]$ as
\begin{align}\label{A_3}
	\setlength\abovedisplayskip{2pt}
	\setlength\belowdisplayskip{2pt}
	&\mathbb{E}[\|o\|^2]\nonumber\\&\!=\mathbb{E}\left[\left\|\nabla{F\left({W}_s\right)}-\nabla{F\left(\tilde{W}_s\right)}+\nabla{F\left(\tilde{W}_s\right)}-g_s\right\|^2\right]\nonumber\\
	&\!\le 2\mathbb{E}\left[ \left\|\frac{1}{K}\sum_{i=1}^IK_i\nabla{F\left(\tilde{W}_s^i\right)}-\nabla{F\left({W}_s\right)}\right\|^2\right]+2\mathbb{E}\left[\left\|\epsilon\right\|^2\right]\nonumber\\
	&\!\le\!\frac{\sum_{i}\!K_i^2\mathbb{E}\!\left[ \left\|\nabla{F\!\left(\tilde{W}_s^i\right)}-\nabla{F\left({W}_s\right)}\right\|^2\right]}{K^2/2I}+2\mathbb{E}\left[\left\|\epsilon\right\|^2\right],
\end{align}
where $\epsilon=\nabla{F\left(\tilde{W}_s\right)}-g_s$. 
Similar to the proof in Appendix A of \cite{chen2021joint}, we have 
\begin{equation}\label{A_4}
	\setlength\abovedisplayskip{2pt}
	\setlength\belowdisplayskip{2pt}
	\mathbb{E}\!\left[\left\|\epsilon\right\|^2\right]\le\frac{4}{K}\mathbb{E}\left(\xi_1+\xi_2\left\|\nabla F\left({W}_s\right)\right\|^2\right)\sum\nolimits_{i=1}^IK_iq_i
\end{equation}
With (\ref{A_3}) and (\ref{A_4}), the $l_2$-norm of gradients is bounded by
\begin{align}\label{A_5}
	%\begin{aligned}
A&\mathbb{E}\left[\|\nabla{F\left({W}_s\right)}\|^2\right]\le\mathbb{E}\left[{F\left(W_s\right)}\right]-\mathbb{E}\left[F\left(W_{s+1}\right)\right]+\nonumber\\&\!\frac{4\xi_1}{\beta K}\!\sum\nolimits_{i=1}^I\!K_iq_i\!+\!\frac{\beta I}{K^2}\!\sum\nolimits_{i=1}^I\!K_i^2\mathbb{E}\!\left[\!\|W_s-\tilde{W}_s^i\|^2\!\right],
	%\end{aligned}
\end{align}
where $A={\left(K-8\xi_2\sum_{i=1}^IK_iq_i\right)}/{2\beta K}$, which should be greater than zero to guarantee convergence.
Because $2\beta\le 1/A \le2\beta/\left(1-8\xi_2\right)$, we substitute the upper bound of $1/A$ into the right hand of (\ref{A_5}):
\begin{align}\label{A_6}
	%\begin{aligned}
		\setlength\abovedisplayskip{2pt}
	    \setlength\belowdisplayskip{2pt}
		\mathbb{E}&\left[\|\nabla{F\left({W}_s\right)}\|^2\right]\le \frac{2\beta\{\mathbb{E}\left[{F\left(W_s\right)}\right]-\mathbb{E}\left[F\left(W_{s+1}\right)\right]\}}{d}\nonumber\\&+\frac{8\xi_1}{dK}\sum_{i}K_iq_i+\frac{2\beta^2 I}{dK^2}\sum_{i}K_i^2\mathbb{E}\left[\|W_s-\tilde{W}_s^i\|^2\right],
	%\end{aligned}
\end{align}
where $d=1-8\xi_2$. Sum up inequalities from $s=0$ to $s=S$, the average $l_2$-norm of gradients is derived as
\begin{align} \label{A_7}
	%\begin{aligned}
	\setlength\abovedisplayskip{2pt}
	\setlength\belowdisplayskip{2pt}
	&\frac{1}{S+1}\sum\nolimits_{s=0}^{S}{\!\mathbb{E} }\left[\left\|\nabla{F\left({W}_s\right)}\right\|^2\right]\overset{\text{(b)}}{\leq} \nonumber \\
	&\!\underbrace{\frac{F\!\left(W_0\right)\!-\!F\!\left(W^*\right)}{d\left(S+1\right)/2\beta}}_{\text{effect of initial model }}\!+\!\underbrace{\frac{8\xi_1}{dK}\!\sum_{i=1}^{I}\!K_i\overline{q}_i}_{\text{effect of packet error}}\!+\!\underbrace{\frac{2\beta^2ID^2}{dK^2}\!\sum_{i=1}^{I}\!K_i^2\overline{\rho}_i}_{\text{effect of network pruning}},
	%\end{aligned}
\end{align}
where $\overline{\rho}_i$ and $\overline{q}_i$ represent the average pruning rate and average packet error rate at UE $i$ during $S+1$ rounds, respectively and $(b)$ stems from $\mathbb{E}[\|W_s-\tilde{W}_s^i\|^2]\le\rho_i\mathbb{E}[\|W_s\|^2]\le\rho_i D^2$ and the fact $F(W^*)\le F(W_{s+1})$.

\bibliography{ref}

\begin{thebibliography}{10}
\providecommand{\url}[1]{#1}
\csname url@samestyle\endcsname
\providecommand{\newblock}{\relax}
\providecommand{\bibinfo}[2]{#2}
\providecommand{\BIBentrySTDinterwordspacing}{\spaceskip=0pt\relax}
\providecommand{\BIBentryALTinterwordstretchfactor}{4}
\providecommand{\BIBentryALTinterwordspacing}{\spaceskip=\fontdimen2\font plus
\BIBentryALTinterwordstretchfactor\fontdimen3\font minus
  \fontdimen4\font\relax}
\providecommand{\BIBforeignlanguage}[2]{{%
\expandafter\ifx\csname l@#1\endcsname\relax
\typeout{** WARNING: IEEEtran.bst: No hyphenation pattern has been}%
\typeout{** loaded for the language `#1'. Using the pattern for}%
\typeout{** the default language instead.}%
\else
\language=\csname l@#1\endcsname
\fi
#2}}
\providecommand{\BIBdecl}{\relax}
\BIBdecl
\renewcommand{\BIBentryALTinterwordstretchfactor}{1}

\bibitem{dinh2021federated}
C.~T. Dinh \emph{et~al.}, ``Federated learning over wireless networks:
  Convergence analysis and resource allocation,'' \emph{IEEE/ACM Trans.
  Networking}, vol.~29, no.~1, pp. 398--409, Feb. 2021.

\bibitem{yang2020energy}
Z.~Yang \emph{et~al.}, ``Energy efficient federated learning over wireless
  communication networks,'' \emph{IEEE Trans. Wireless Commun.}, vol.~20,
  no.~3, pp. 1935--1949, Mar. 2021.

\bibitem{luo2021cost}
B.~Luo \emph{et~al.}, ``Cost-effective federated learning design,'' in
  \emph{Proc. IEEE INFOCOM}, Vancouver, Canada, May 2021, pp. 1--10.

\bibitem{xu2021client}
J.~Xu \emph{et~al.}, ``Client selection and bandwidth allocation in wireless
  federated learning networks: A long-term perspective,'' \emph{IEEE Trans.
  Wireless Commun.}, vol.~20, no.~2, pp. 1188--1200, Feb. 2021.

\bibitem{chen2021joint}
M.~Chen \emph{et~al.}, ``A joint learning and communications framework for
  federated learning over wireless networks,'' \emph{IEEE Trans. Wireless
  Commun.}, vol.~20, no.~1, pp. 269--283, Jan. 2021.

\bibitem{ren2021joint}
J.~Ren \emph{et~al.}, ``Joint resource allocation for efficient federated
  learning in internet of things supported by edge computing,'' in \emph{Proc.
  ICC Workshops}, Montreal, Canada, Jun. 2021, pp. 1--6.

\bibitem{molchanov2019importance}
P.~Molchanov \emph{et~al.}, ``Importance estimation for neural network
  pruning,'' in \emph{Proc. CVPR}, California, USA, Jun. 2019, pp.
  11\,264--11\,272.

\bibitem{li2021talk}
L.~Li \emph{et~al.}, ``To talk or to work: Flexible communication compression
  for energy efficient federated learning over heterogeneous mobile edge
  devices,'' in \emph{Proc. IEEE INFOCOM}, Vancouver, Canada, May 2021, pp.
  1--10.

\bibitem{jiang2020model}
\BIBentryALTinterwordspacing
Y.~Jiang \emph{et~al.}, ``Model pruning enables efficient federated learning on
  edge devices,'' Oct. 2020. [Online]. Available:
  \url{https://arxiv.org/abs/1909.12326}
\BIBentrySTDinterwordspacing

\bibitem{liu2021adaptive}
S.~Liu \emph{et~al.}, ``Adaptive network pruning for wireless federated
  learning,'' \emph{IEEE Wireless Commun. Lett.}, vol.~10, no.~7, pp.
  1572--1576, Jul. 2021.

\bibitem{Xi2011ageneral}
Y.~Xi \emph{et~al.}, ``A general upper bound to evaluate packet error rate over
  quasi-static fading channels,'' \emph{IEEE Trans. Wireless Commun.}, vol.~10,
  no.~5, pp. 1373--1377, May 2011.

\end{thebibliography}
\bibliographystyle{IEEEtran}

\end{document}